\documentclass{article}
\usepackage{spconf,amsmath,graphicx}
\usepackage{multirow,tikz,pgfplots}
\usepackage{color, colortbl}
\definecolor{Gray}{gray}{0.9}

\title{Information-Weighted Neural Cache Language Models for ASR}

\name{Lyan Verwimp$^1$\thanks{This research is funded by the Flemish government agency IWT (project 130041, SCATE).}, Joris Pelemans$^{2,*}$\thanks{$^*$This work was done while the researcher was a member of ESAT -- PSI, KU Leuven and does not contain Apple proprietary information.}, Hugo Van hamme$^1$, Patrick Wambacq$^1$}
\address{$^1$ESAT -- PSI, KU Leuven, Leuven, Belgium \\
$^2$Apple Inc., Cupertino, USA}

\date{}

\begin{document}
\maketitle
\renewcommand{\thefootnote}{\arabic{footnote}}
\begin{abstract}
Neural cache language models (LMs) extend the idea of regular cache language models by making the cache probability dependent on the similarity between the current context and the context of the words in the cache. We make an extensive comparison of `regular' cache models with neural cache models, both in terms of perplexity and WER after rescoring first-pass ASR results. Furthermore, we propose two extensions to this neural cache model that make use of the content value/information weight of the word: firstly, combining the cache probability and LM probability with an information-weighted interpolation and secondly, selectively adding only content words to the cache. We obtain a 29.9\%/32.1\% (validation/test set) relative improvement in perplexity with respect to a baseline LSTM LM on the WikiText-2 dataset, outperforming previous work on neural cache LMs. Additionally, we observe significant WER reductions with respect to the baseline model on the WSJ ASR task.
\end{abstract}

\section{Introduction}

Language models (LMs) play a crucial role in many speech and language processing tasks such as speech recognition and machine translation. The current state of the art are recurrent neural network (RNN) based LMs \cite{Mikolov10}, and more specifically long short-term memory (LSTM)~\cite{lstm} LMs~\cite{lstm_lm,Joz}. 

This work focuses on extending LSTM LMs with a cache~\cite{cache}, which gives the LM two main advantages: firstly, the cache is a simple and computationally cheap manner to adapt the LM to the current topic since it increases the probabilities of words recently seen. Secondly, Kuhn and De Mori~\cite{cache} first proposed a cache model as a means to overcome one of the limitations of n-gram LMs, namely that they are not capable of modeling short-term patterns in word use. 
One can argue that LSTMs are better at this because they should be able to remember words for a longer time. However, the effect of a single word might be obscured after a certain amount of time steps since with each time step, a complete update of the cell state and hidden state is done. Explicitly adding previously seen words to a cache and combining the cache and LSTM probabilities can counteract this effect. 


Grave et al.~\cite{Grave} recently proposed a \textit{continuous} cache (henceforth referred to as \textit{neural} cache, as opposed to Kuhn and De Mori's \textit{regular} cache~\cite{cache}), that assigns cache probabilities based on the similarity between the current hidden state and the hidden state associated with the word in the cache. As a result, words previously seen in a similar context get a higher cache probability than words seen in a different context. The advantages of the neural cache for neural LMs have also been demonstrated by Merity et al.~\cite{Merity} and Khandelwal et al.~\cite{Khandelwal}, but they use only a neural cache and do not compare with a regular cache, and only evaluate on textual data.
%
%

The manner in which the cache probability is combined with the LSTM LM probability is a crucial part in cache LMs. Grave et al.~\cite{Grave} mention linear interpolation and another method called global normalization, but the latter approach does not outperform linear interpolation. One can argue that linear interpolation -- assigning the same weight to the cache probability regardless of the type of word -- is suboptimal since the cache probability will be more helpful for content words whereas the LM probability should receive more weight for function words.
Kuhn and De Mori~\cite{cache} train separate interpolation weights and use separate caches for each part-of-speech (POS), combining the cache model with a class-based n-gram model in which the classes correspond to POS classes. However, we believe that keeping a separate cache for certain classes such as determiners and prepositions does not make much sense because for those classes, the overall statistics captured by the LM should be much more meaningful. For example, having seen the word \textit{in} multiple times before does not increase the probability of seeing it again. On the other hand, certain constructions such as verbs requiring this preposition (e.g. \textit{believe, specialize}) or sentences in which adverbial phrases indicating time or space are plausible (e.g. \textit{On the day of its release \textbf{in} Japan, Work began on the Tower Building \textbf{in} 1840}\footnote{Extracts from WikiText~\cite{Wiki}.}), do increase the probability of seeing \textit{in}, and such dependencies should be captured by the LM.
Another disadvantage of Kuhn and De Mori's approach is that it needs LM training data that has been tagged for POS. 

The goal of this paper is two-fold: 1) we propose to weigh the cache model with a measure that is solely based on the frequency statistics of the training data and which hence does not need training data augmented with POS: the \textit{information weight} (IW) of a word (see section~\ref{interp} for a definition). This measure can be automatically calculated, and we investigate whether it can give further improvements for a cache model.
The IW can be used in \textit{information-weighted interpolation} and/or for an \textit{information-weighted selective} cache, to which only words with an IW larger than a certain threshold are added. 

Secondly, as far as we know no comparison between regular and neural cache models for automatic speech recognition (ASR), an application for which cache models typically give good improvements, has been made. Thus, 2) the second goal of this paper is to investigate what advantage a (information-weighted) neural cache can have for ASR. 
Typically, neural LMs for ASR are used in a multi-pass approach: in a first pass, the search space is narrowed down with the help of a simple (usually n-gram or FST) LM, producing a set of hypotheses in the format of a lattice or an N-best list. We evaluate our models by rescoring N-best lists.

In the remainder of this paper, we will first define the different cache models and different manners of interpolation (section~\ref{cache}) and validate the proposed methods experimentally (section~\ref{exp}). We end with a conclusion and an outlook to future work (section~\ref{concl}). 

\section{Cache language models}
\label{cache}



\subsection{Regular cache}
\label{sec:reg-cache}

A regular cache model keeps track of the words previously seen and computes the probability of the next word being $w_t$ as follows:
%
%
%
%

\begin{equation}
\label{reg-cache}
P_{c}(w_{t}~|~w_{t-|C|-1}\ldots~w_{t-1}) = \frac{\sum\limits_{j = t-|C|-1}^{t-1} I_{\{w_j=w_t\}}}{|C|}
%
%
%
%
%
%
\end{equation}
%
%
where $C$ is the cache and $|C|$ its size, $t$ is the current time step and $I$ the indicator function, which is 1 if $w_j = w_t$ and 0 otherwise. An exponential decay can be used to give more weight to words seen more recently: $I_{\{w_j = w_t\}} e^{-\alpha~(t-j)}$, with $\alpha$ the decay rate.


\subsection{Neural cache}

The neural cache model proposed by Grave et al.~\cite{Grave} calculates the cache probability as follows:

\begin{equation}
\begin{aligned}
\label{neur-cache}
\begin{split}
P_{c}(w_{t}~&|~(w_{t-|C|-1},h_{t-|C|-2})\ldots~(w_{t-1},h_{t-2})) = \\
&\frac{\sum\limits_{j=t-|C|-1}^{t-1} I_{\{w_{j} = w_t\}} e^{(\theta~h_t^T~h_{j-1})}}{\sum\limits_{w_i \in C}~\sum\limits_{j=t-|C|-1}^{t-1} I_{\{w_{j} = w_i\}} e^{(\theta~h_i^T~h_{j-1})}}
\end{split}
\end{aligned}
\end{equation}
%
%
%
%
%
%
%
%
In the equation above, $h_{j-1}$ is the hidden state of the LSTM cell, which corresponds to the cell's output that is used to predict the next word $w_j$. In other words, the cache $h_c$ contains pairs of hidden states and corresponding target words. $\theta$ is a hyperparameter that controls the flatness of the distribution -- the lower $\theta$, the flatter the distribution of the cache probabilities. Given that $h_t^T~h_{j-1}$ is in the range (-1,1), the minimum value of $e^{(\theta~h_t^T~h_{j-1})}$ will always be $< 1$ and the maximum value $> 1$ (for $\theta > 0$), as opposed to a regular cache, where the value in the denominator is always 1.
%
%
%
%

If a word is in the cache, its cache probability is calculated based on the inner product between the current hidden state of the LSTM $h_t$ and the hidden state stored in the cache $h_{j-1}$. Hence, the probability is not solely based on whether the word has been seen before or not -- as happens in regular cache models --, but also based on the similarity between the current context and the context in which the cache word occurred, since we assume that the hidden state of the LSTM captures the previous context of the word. For example, in a Wikipedia article about the Chinese poet \textit{Du Fu}, we see that the neural cache assigns a much higher probability (compared to the unigram probability of the regular cache) to \textit{Fu} if \textit{Du} is the input word, since these two words occur together very frequently.

\subsection{Information-weighted interpolation of probabilities}
\label{interp}

Grave et al.~\cite{Grave} describe two ways of combining the cache probability with the standard probabilities output by the network, of which linear interpolation performs slightly better. 
%
%


However, knowing that a word has been seen before might be more meaningful for certain words than for others. 
Hence, we propose to exploit this fact by assigning a larger interpolation weight to words with a large content value than to words with a low content value, such as function words. The information weight $\lambda_i$ of a word is defined as follows:
%
%
%
%
\begin{equation}
\label{weight}
\lambda_i = 1 + \frac{\sum_{j=1}^{N}~P(j|i)~\log(P(j|i))}{\log(N)}
\end{equation}
where $P(j|i)$ is the probability of document $j$ given that word $i$ occurs in it, and is equal to the  frequency of word $i$ in document $j$ divided by the frequency of word $i$ in the whole corpus. $N$ is the number of documents in the corpus; for our experiments we split the corpus in documents of an equal amount of sentences. Equation~\ref{weight} corresponds to 1 -- normalized entropy, hence giving higher importance to words that are sparsely distributed across the corpus. This measure is among others used by Dumais~\cite{dumais} and Bellegarda~\cite{Bellegarda} as a global weighting for the Latent Semantic Analysis word-document matrix. The values for $\lambda$ range between 0, for words that are uniformly distributed among the corpus, and 1, for words that appear in only 1 document. For example, \textit{bilingual} has a weight of $\lambda = 0.910$ whereas \textit{the} has a weight of $\lambda = 0.019$ for the WikiText-2 corpus~\cite{Wiki}. 

The final probability for word $i$ is obtained by information-weighted linear interpolation:

\begin{equation}
\label{info-linear}
\begin{split}
&P(w_t~|~history) = \\
&\frac{(1 - \gamma~\lambda_t) P_{LM}(w_t~|~h_t) + (\gamma~\lambda_t) P_{c}(w_t~|~h_c)}{\sum\limits_{w \in V} (1 - \gamma~\lambda_t) P_{LM}(w_t~|~h_t) + (\gamma~\lambda_t) P_{c}(w_t~|~h_c)}
\end{split}
\end{equation}
where $\gamma$ is a hyperparameter that scales the information weights and needs to be set empirically. We only consider values smaller than or equal to 0.5, such that the cache probability will never be assigned a higher weight than the LM probability. Thus, the LM probabilities have more weight for frequent words, while for topical words a combination of LM and cache probabilities is used, which will lead to increased probabilities for topical words in the cache.



\subsection{Information-weighted selective cache}
\label{select}

Another manner to use the information weight of a word is to use it as a hard threshold: only words with an information weight greater than or equal to a certain threshold $\phi$ are added to the cache. This approach has the advantage that no space is `wasted' on words with low content value. It effectively enlarges the scope of the cache. A selective cache can be used in combination both with a regular and neural cache model, and with linear and information-weighted interpolation.

\begin{table*}[t]
\centering
\begin{tabular}{llll|c|c|c|c|c||c|c|c|c|c}
\hline
&\multirow{2}{*}{}&&&\multicolumn{5}{c||}{\textbf{Regular cache}}&\multicolumn{5}{c}{\textbf{Neural cache}}\\
\hline
\textbf{Model}&\textbf{Size}&\textbf{Interp.}&\textbf{IW select.}&\textbf{Valid}&\textbf{Test}&$\lambda~/~\gamma$&$\phi$&$\alpha$&\textbf{Valid}&\textbf{Test}&$\lambda~/~\gamma$&$\phi$&$\theta$ \\
\hline 
\hline
Baseline~\cite{Grave}&-&-&-&-&\cellcolor{Gray}99.3&-&-&-&-&\cellcolor{Gray}99.3&-&-&- \\
Cache~\cite{Grave}&100&linear&-&-&\cellcolor{Gray}-&-&-&-&-&\cellcolor{Gray}81.6&-&-&- \\
\hline
Our baseline&-&-&-&102.5&\cellcolor{Gray}97.6&-&-&-&102.5&\cellcolor{Gray}97.6&-&-&- \\
Cache&100&linear&no&92.6&\cellcolor{Gray}87.9&0.10&-&-&85.6&\cellcolor{Gray}81.1&0.10&-&0.3 \\
Cache&100&IW&no&91.9&\cellcolor{Gray}87.1&0.20&-&-&86.1&\cellcolor{Gray}81.9&0.25&-&0.3 \\
\hline
Our baseline&-&-&-&103.0&\cellcolor{Gray}98.6&-&-&-&103.0&\cellcolor{Gray}98.6&-&-&- \\
Cache&100&linear&yes&90.7&\cellcolor{Gray}86.8&0.05&0.2&-&84.4&\cellcolor{Gray}81.3&0.05&0.2&0.3 \\
Cache&100&IW&yes&\underline{86.3}&\cellcolor{Gray}\underline{82.6}&0.25&0.4&-&\underline{79.6}&\cellcolor{Gray}\underline{76.7}&0.35&0.2&0.3 \\
\hline
\hline
Baseline~\cite{Grave}&-&-&-&-&\cellcolor{Gray}99.3&-&-&-&-&\cellcolor{Gray}99.3&-&-&- \\
Cache~\cite{Grave}&2000&linear&-&-&\cellcolor{Gray}-&-&-&-&-&\cellcolor{Gray}68.9&-&-&- \\
\hline
Our baseline&-&-&-&103.4&\cellcolor{Gray}98.1&-&-&-&103.4&\cellcolor{Gray}98.1&-&-&- \\
Cache&2000&linear&no&88.6&\cellcolor{Gray}83.8&0.10&-&0.005&73.4&\cellcolor{Gray}69.6&0.15&-&0.3 \\
Cache&2000&IW&no&87.4&\cellcolor{Gray}83.0&0.30&-&0.004&73.9&\cellcolor{Gray}70.9&0.35&-&0.3\\
\hline
Our baseline&-&-&-&103.3&\cellcolor{Gray}97.5&-&-&-&103.3&\cellcolor{Gray}97.5&-&-&- \\
Cache&2000&linear&yes&88.3&\cellcolor{Gray}84.0&0.05&0.2&0.008&76.6&\cellcolor{Gray}73.4&0.20&0.1&0.3 \\
Cache&2000&IW&yes&\textbf{82.4}&\cellcolor{Gray}\textbf{78.3}&0.35&0.2&0.006&\textbf{72.4}&\cellcolor{Gray}\textbf{66.2}&0.45&0.2&0.3 \\
\hline
\end{tabular}
\caption{Validation and test perplexities for LSTM LMs with a cache trained on WikiText-2, along with their optimal hyperparameters (if applicable). The column $\lambda~/~\gamma$ contains the optimal interpolation weight $\lambda$ for models with linear interpolation and the optimal scale for the IW $\gamma$ for models with IW interpolation. }
\label{tab:all-cache}
\vspace{-1em}
\end{table*}

\section{Experiments}
\label{exp}

\subsection{Setup}

For perplexity experiments, we train our LMs on WikiText-2~\cite{Wiki}. This dataset is also used by Grave et al.~\cite{Grave} and contains 2M words for training, 220k words for validation, 250k words for testing and has a vocabulary of 33k words.

All training and perplexity experiments are implemented with TensorFlow~\cite{tf}.
The LSTM LMs trained on WikiText have 1 layer of 512 LSTM cells, randomly initialized with a uniform distribution between -0.05 and 0.05. We use mini-batches of 20 samples and train the LSTMs with backpropagation through time (35 steps) and 50\% dropout~\cite{dropout}. The norm of the gradients is clipped at 5. We train with stochastic gradient descent: during the first 6 epochs, the learning rate is 1, after which we apply an exponential decay: the learning rate for epoch $i$, $\eta_{i}$, is equal to $\alpha~\eta_{i-1}$, with $\alpha$ the learning rate decay which is set to 0.8. After each epoch, we check how often the validation perplexity has not improved with respect to the previous epochs: if it has not improved for 3 times, we stop training. If the validation perplexity keeps on improving, we stop training anyway after 39 epochs. We train on discourse level, which means that we do not reset the LSTM state at sentence boundaries and predict the first word of the next sentence from the end-of-sentence token (for more information, see Verwimp et al.~\cite{tf-lm}). This model gives a baseline perplexity close the the one reported by Grave et al.~\cite{Grave}.
The hyperparameters specific to the cache model are optimized on the validation set. We provide all values for the hyperparameters together with the results, since it has been shown that these have an important influence~\cite{melis}.  

We use Kaldi~\cite{kaldi} for the ASR experiments. We follow the standard \textit{nnet2} recipe for Wall Street Journal (WSJ), generating 1000-best lists for rescoring. The first pass of the recognition makes use of a 3-gram LM with a vocabulary of 120k words. The training data for the LSTM LMs consists of 37M words with a vocabulary of 40k. It contains years 87--89 with non-verbalized punctuation. The new LM probability is calculated as follows:

\begin{equation}
\label{p-new}
\begin{split}
P_{new}(W) &= (1 - \lambda_{LSTM})*P_{ngram}(W) \\
&+ \lambda_{LSTM}*P_{LSTM+C}(W)
\end{split}
\end{equation}
where $P_{ngram}$ is the first-pass LM probability and $P_{LSTM+C}$ is equal to the probability in equation~\ref{info-linear}. A scaled version of $P_{new}(W)$ ($\beta$ = scaling factor) is combined with the score of the acoustic model $P(O|W)$:

\begin{equation}
\label{rescore}
\begin{split}
\log P(W) = \log P(O|W) + \beta~\log (P_{new}(W)) + L ~\omega
\end{split}
\end{equation}
$\omega$ is the word insertion penalty and $L$ the number of words in the hypothesis.
For every model type, hyperparameters such as word insertion penalty $\omega$, LM scale $\beta$, LSTM LM weight $\lambda_{LSTM}$ are optimized on the validation set. 
The LSTM LMs used for rescoring are trained with TensorFlow~\cite{tf} with the same hyperparameters as the WikiText models, except that we train on sentence level instead of discourse level.

We use the bootstrap method of Bisani and Ney~\cite{bisani} for significance analysis of the WER results, which is implemented as Kaldi's \textit{compute-wer-bootci} script. This method generates (by default 10k) random bootstrap samples of the data and calculates the percentage of times the bootstrap sample from system B improves over the bootstrap sample from system A: the result is a `probability of improvement' (poi) of system B over system A.

Since the length of the cache is usually longer than the length of the hypotheses in a single N-best list, we would like to be able to transfer the cache across lists. We keep track of the probabilities and caches for every hypothesis in the list, and initialize the cache of the next N-best list with the cache of the most likely hypothesis from the previous list. 

\subsection{Perplexity results}

We report the perplexity of the LSTM LMs in table~\ref{tab:all-cache}. Due to implementation requirements, we have to train 4 different models: one with a non-selective cache of size 100, one with a selective cache of size 100 and the same options for a cache size of 2000. The baseline perplexities for LSTMs without cache of these four models do not differ much, but we report them all for the sake of completeness (`Our baseline' in the table). All perplexity results shown below the same baseline can thus be attributed solely to the cache component since they are obtained with the same trained model.

Firstly, we combine the LSTM with a regular cache model with linear interpolation (left part of the table) and see that this simple combination already gives us improvements: 11.4\%/11.6\% relatively on the validation/test set for a cache size of 2000 (with exponential decay). This result confirms our hypothesis, namely that an LSTM in itself is not sufficient in remembering the relevant words in the previous context. Using information-weighted (\textit{IW}) interpolation or IW selective cache usually gives additional improvements. The two IW-based techniques seem to give complimentary improvements, because the combination yields the best perplexity result (relative improvement of 20.2\%/19.7\%) for the regular cache model. 


The results for LSTM LMs with a neural cache are reported in the right part of the table. 
We observe that, similarly to the regular cache, the combination of both information-based extensions yields perplexity improvements, also with respect to the results reported by Grave et al.~\cite{Grave}. Using only one of the two IW-based techniques does not result in better perplexity results.
Additionally, we observe that the neural cache consistently outperforms the regular cache. 
The best result is obtained with a neural selective cache of size 2000 with information-weighted interpolation: a relative improvement of 29.9\%/32.1\% (validation/test) with respect to our own baseline.

\subsection{Analysis}

Taking a closer look at the optimal hyperparameters, we firstly observe that $\gamma$ is larger for the neural cache than for the regular cache. For the best model, $\gamma = 0.45$, resulting in weights close to 0.5 for words with a large IW $\lambda$, which indicates a large reliance on the cache for content words. 

Secondly, the optimal threshold $\phi$ for most IW selective cache models is quite low, 0.2. This threshold essentially excludes only 222 word types from the cache, that albeit form more than 50\% of the test set (135k out of 240k). 

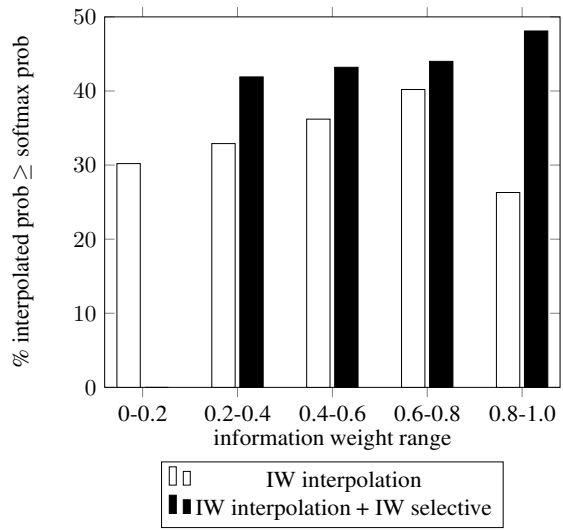
\begin{figure}[ht]
\centering
\resizebox{7.5cm}{7cm}{\begin{tikzpicture}
\begin{axis}[
	symbolic x coords={0-0.2, 0.2-0.4, 0.4-0.6, 0.6-0.8, 0.8-1.0},
    ylabel = {\% interpolated prob $\geq$ softmax prob},
    ybar,
    legend style={at={(0.5,-0.2)}, anchor=north},
    xlabel = {information weight range},
    ymax=50,ymin=0]
\addplot[fill=white] coordinates { (0-0.2, 30.2) (0.2-0.4, 32.9) (0.4-0.6, 36.2) (0.6-0.8, 40.2) (0.8-1.0, 26.3) };
\addplot[fill=black] coordinates { (0-0.2, 0) (0.2-0.4, 41.9) (0.4-0.6, 43.2) (0.6-0.8, 44.0) (0.8-1.0, 48.1) };
\legend{IW interpolation, IW interpolation + IW selective}
\end{axis}
\end{tikzpicture}}
\caption{The proportion of times that the interpolated probability is higher than or equal to the baseline LM probability with respect to several ranges of IWs, for the best neural cache models with IW interpolation and with the combination of IW selective cache and IW interpolation.}
\label{fig:data_distr}
\vspace{-1em}
\end{figure}

In figure~\ref{fig:data_distr}, we plot how often the information-thresholded cache assigns a larger interpolated probability to the target word than the original LM probability.
Firstly, since $\phi$ is 0.2 for the selective model, very frequent low-content words (with IW $< 0.2$) will always have a lower interpolated probability than the softmax probability because the cache probability is 0. However, for IWs larger than 0.2, the selective model is more likely to result in a better interpolated probability, and that probability increases if the IW itself increases. 

As a result, if we look at the predictions over all IWs, the non-selective model outperforms the LM probability in 31.9\% of the cases and the selective model only in 25.1\% of the cases. But for IWs $> 0.2$, the selective models outperforms the baseline in 43.5\% of the cases, and the non-selective one only in 33.9\% of the cases. 
When the interpolated probability is higher than the softmax probability, the relative improvement of the interpolated probability with respect to the softmax probability, averaged over the test set, is 55.4\% for the non-selective model and 54.3\% for the selective model. 
For IWs $> 0.2$, the relative improvement is higher for the selective model (67.8\%) than for the non-selective one (65.8\%), although the difference is small.
Thus, the cache model combining the IW interpolation and the IW selection gives worse estimates for very frequent words than the baseline model and the cache with only IW interpolation since they have a cache probability of 0, but better estimates for topical words, resulting in an overall perplexity improvement.

\subsection{Speech recognition results}

In table~\ref{tab:asr-wsj}, we report the perplexity results on the WSJ validation set and the validation and test WER after rescoring with LSTM LMs. Similarly as for WikiText, we report the results for two different baseline models, a selective one and a non-selective one. Notice that all results within the upper part and respectively the lower part of the table are obtained with the same trained model. Thus, differences in performance can be attributed to the cache parts solely and not to different initialization settings that might lead to better convergence of the model during training. 
As a sanity check, we tested whether the differences in WER between the two baseline models were significant: according to the bootstrap method~\cite{bisani}, the probability of the first baseline improving over the second baseline is 39.97.5\%/73.57\% (validation/test set). Thus, we conclude that the difference between the baselines is not significant.
We only test models with a cache size of 100, since preliminary perplexity results with a larger cache size did not give substantial improvements with respect to a cache of 100 for this dataset.

Firstly, we observe that the differences in perplexity are much smaller than for WikiText: this shows that on the WSJ dataset, much less can be gained by using a cache. This is not surprising given the fact that WikiText is specifically designed to contain many long-term dependencies~\cite{Wiki}. Secondly, the neural cache is not consistently better than the regular cache, as opposed to WikiText. The combination of a neural cache with linear interpolation gives the best results, both for a non-selective and a selective model, but the improvement is only 4.6\%/4.1\% (non-selective/selective) relative to the baseline.  
Overall, the differences in perplexity between the cache models are quite small, but notice that this dataset is quite small (8k words, compared to 210k/240k words for the validation/test set of WikiText) and hence we should be careful in drawing strong conclusions based on these results.

The cache models give WER improvements between 0.2 and 0.32 absolute for rescoring the 1000-best lists of the validation set. Between brackets we indicate the probability of improvement with respect to the baseline: all results except the neural cache with linear interpolation are significant with a confidence larger than 95\%, but the best results are still quite far from the oracle WER of 2.18. We observe that the best model in terms of perplexity does not yield the best WER results, and that the WER differences between the cache models are not significant.

\begin{table*}[t]
\centering
\begin{tabular}{l|c|c|c|c|c|c|c|c}
\hline
\textbf{Model}&\textbf{Valid PPL}&\textbf{Valid WER (poi)}&\textbf{Test WER (poi)}&$\lambda$ / $\gamma$&$\phi$&$\lambda_{LSTM}$&$\beta$&$\omega$ \\
\hline
baseline&176.9&5.61&2.55&-&-&0.5&18&0 \\
regular linear&170.5&\textbf{5.36} (96.55)&2.34 (93.95)&0.05&-&0.5&15&-0.5 \\
regular IW&171.2&5.41 (96.47)&\textbf{2.25} (\textbf{99.34})&0.4&-&0.5&13&-0.5 \\ 
neural linear&\textbf{168.8}&5.38 (94.63)&2.43 (80.66)&0.05&-&0.5&15&-0.5 \\
neural IW&170.3&5.38 (\textbf{97.21})&2.39 (89.87)&0.7&-&0.4&13&0 \\ 
\hline
baseline&175.1&5.59&2.64&-&-&0.5&18&0\\
select. reg. linear&170.2&5.32 (99.10)&2.68 (33.68)&0.05&0.05&0.3&13&0\\ 
select. reg. IW&168.1&\textbf{5.27} (\textbf{99.8}1)&\textbf{2.61} (\textbf{58.75})&0.1&0.05&0.3&13&0\\ 
select. neur. linear&\textbf{167.9}&5.32 (99.10)&2.64 (46.33)&0.05&0.05&0.3&13&0 \\
select. neur. IW&168.8&5.31 (99.57)&2.63 (52.32)&0.1&0.005&0.3&13&0 \\
\hline
\end{tabular}
\caption{Perplexities on the validation set of WSJ and WERs after rescoring the 1000-best lists of the validation and test set. All cache models have a cache size of 100. `poi' = probability of improvement~\cite{bisani} of the cache model with respect to its baseline. The 5 last columns contain the hyperparameter values optimized on the validation set by rescoring, which are also used for rescoring the test set. The column $\lambda~/~\gamma$ contains the optimal $\lambda$ for models with linear interpolation and the optimal $\gamma$ for models with IW interpolation. We do not include the decay values since exponential decay did not improve the results. Oracle WER for the 1000-best lists is 2.18 (validation) and 1.21 (test). }
\label{tab:asr-wsj}
\vspace{-1em}
\end{table*}

On the test set, the non-selective models achieve WER reductions between 0.12 and 0.3 absolute; notice that the largest possible reduction is 1.43 (oracle WER 1.21). However, only the regular cache with IW interpolation achieves a significant WER reduction with respect to the baseline.
This model also significantly improves the results for the neural models (both with a confidence $>$ 98\%), but not significantly the result of the regular cache with linear interpolation (poi of 89.63\%). The regular cache with linear interpolation significantly improves its neural counterpart (poi 98.36\%), but not the neural cache with IW interpolation (poi 72.26\%). Given that there are no significant differences between the cache models on the validation set on which the hyperparameters are optimized, this suggests that the regular cache models are more robust against hyperparameter settings that are possibly not optimal for the test set. 

The selective models on the other hand, show no or only minor improvement with respect to their baseline model on the test set. The best model only gives 0.03 absolute reduction in WER. We investigated several possible explanations for this large discrepancy between the results of the selective models on the validation and on the test set. 
Firstly, it is possible that the selective models are much more sensitive to hyperparameter settings, but tuning the hyperparameters on the test set as an oracle experiment does not improve the results.
A second possible explanation is that the selective models only work well on data with a specific structure/specific patterns, and both WikiText and the validation set have such structure while the test set does not. However, if we calculate perplexity of the test set, the selective neural cache with IW interpolation is among the best models (5.7\% relative improvement). On the other hand, we observe with this dataset that perplexity results do not always extrapolate nicely to WER results. This might be because the perplexity is calculated on the references only, while the WER is the result of rescoring many hypotheses, of which some might be more probable than the reference according to the LM. If the wrong hypothesis is selected, the wrong cache is transferred across N-best lists. Since the selective models have a larger scope because no space is wasted on function words, the wrong words are effectively longer in the cache. The effect of wrong words in the cache might be worse for certain datasets (e.g. the WSJ test set) than for others (e.g. the WSJ validation set).
A very crude measure that provides some evidence for this explanation of the validation versus test set discrepancy, is counting the number of times that a wrongly selected hypothesis is followed by another wrongly selected hypothesis. We observe that for all non-selective models, this number is lower for the cache models than for the baseline model, while for all selective models, this number is higher for the cache models than for the baseline model, providing plausible evidence for the fact that errors are propagated longer for selective models.
This is even the case for the validation set, but on this dataset, the gains that the selective model provides are larger than the losses, such that the overall WER is lower for the cache models than for the baseline model.

\section{Conclusion and future work}
\label{concl}

In this work, we present an extensive evaluation and comparison of regular and neural cache LMs, in terms of perplexity and WER. Additionally, we propose two extensions based on the information weight/content value of words that can be applied to any cache model: IW interpolation of cache and LM probabilities and an IW selective cache.

On the WikiText benchmark, we obtain large improvements in perplexity, the largest one obtained by the combination of the two information-weighted extensions proposed in this paper. 
Additionally, we observe that neural cache models consistently outperform regular cache models on this dataset.

We also test the cache models on ASR by rescoring 1000-best lists of WSJ. 
We observe significant improvements with respect to a baseline LM on the validation set, while on the test set the regular cache with IW interpolation was the only model giving a significant improvement. As opposed to WikiText, neural cache models are not consistently better than regular cache models, and the combination of the two information-weighted extensions is not the best option. The neural cache seems to be more sensitive to hyperparameter settings, since the regular cache models are significantly better than the neural cache models on the test set, but not on the validation set. The selective models provide significant gains on the validation set, but not on the test set: we hypothesize that those models are more sensitive to the selection of wrong hypotheses in previous segments, because the wrong words stay longer in the cache.

In the future, we would like to test our models on ASR tasks that might give more gains, and test a `smarter' manner of transferring the cache across N-best lists, e.g.~a weighted combination of all hypotheses from the previous list.



\bibliography{slt}
\bibliographystyle{IEEEbib}

\appendix

\end{document}